\documentclass[runningheads]{llncs}
\usepackage[pdftex]{graphicx}

\usepackage{tikz}
\usepackage{comment}
\usepackage{bm}
\usepackage{booktabs} 
\usepackage{amsmath,amssymb} 
\usepackage{color}
\usepackage{hyperref}
\usepackage{multirow}
\usepackage{subfigure}
\usepackage{pdfpages}
\usepackage[font=small,labelfont=bf]{caption}

\usepackage[accsupp]{axessibility}  

\begin{document}

\pagestyle{headings}
\mainmatter
\def\ECCVSubNumber{541}  

\title{Cross-modal Prototype Driven Network for Radiology Report Generation} 


%
\author{Jun Wang \and
Abhir Bhalerao \and
Yulan He}
\authorrunning{Wang et al.}
%
\institute{Department of Computer Science, University of Warwick, UK\\
\email{\{jun.wang.3, abhir.bhalerao, yulan.he\}@warwick.ac.uk}}
\maketitle
\begin{abstract}
Radiology report generation (RRG) aims to describe automatically a radiology image with human-like language and could potentially support the work of radiologists, reducing the burden of manual reporting. Previous approaches often adopt an encoder-decoder architecture and focus on single-modal feature learning, while few studies explore cross-modal feature interaction. Here we propose a Cross-modal PROtotype driven NETwork (XPRONET) to promote  cross-modal pattern learning and exploit it to improve the task of radiology report generation. This is achieved by three well-designed, fully differentiable and complementary modules: a shared cross-modal prototype matrix to record the cross-modal prototypes; a cross-modal prototype network to learn the cross-modal prototypes and embed the cross-modal information into the visual and textual features; and an improved multi-label contrastive loss to enable and enhance  multi-label prototype learning. XPRONET obtains substantial improvements on the IU-Xray and MIMIC-CXR benchmarks, where its performance exceeds recent state-of-the-art approaches by a large margin on IU-Xray and comparable performance on  MIMIC-CXR.\footnote{The code is publicly available at~\href{https://github.com/Markin-Wang/XProNet}{https://github.com/Markin-Wang/XProNet}}
\keywords{Radiology Report Generation, Cross-Modal Pattern Learning, Prototype Learning, Transformers}
\end{abstract}

\section{Introduction}
Radiology images, e.g., X-Ray and MRI, are widely used in medicine to support disease diagnosis. Nonetheless, traditional clinical practice is laborious since it requires the medical expert, such as a radiologist,  to carefully analyze an image and then produce a medical report, which often takes more than five minutes~\cite{jing2018automatic}. This process could also be error-prone due to subjective factors, such as fatigue and distraction. Automatic radiology report generation, as an alternative to expert diagnosis, has therefore gained increasing attention from researchers. Automatic medical report generation has the potential to rapidly produce a report and assist a radiologist to make the final diagnosis significantly reducing the workload of radiologists and saving medical resources, especially in developing countries where well-trained radiologists can be in short supply.

Owing to developments in computer vision models for image captioning and availability of large-scale datasets, recently there have been significant advancements in automated radiology report generation~\cite{zhang2020radiology,jing2018automatic,liu2019clinically}. Nevertheless, radiology report generation 
still remains  a challenging task and is far from being solved. The reasons are three-fold. Firstly, unlike the traditional image captioning task which often produces only a single sentence, a medical report consists of several sentences and its length might be four-times longer than an image caption. Secondly, medical reports often exhibit more sophisticated linguistic and semantic patterns. 
Lastly, commonly used datasets suffer from notable data biases: the majority of the training samples are of normal cases, any abnormal regions often only exist in a small parts of an image, and even in pathological cases, most statements may be associated with a description of normal findings, e.g. see~\autoref{fig:example_vis}. Overall, these  problems present a substantial challenge to the modelling of cross-modal pattern interactions and learning informative features for accurate report generation.
\begin{figure}[t]
\centering
\begin{minipage}{.55\textwidth}
  \centering
  \includegraphics[width=.95\linewidth]{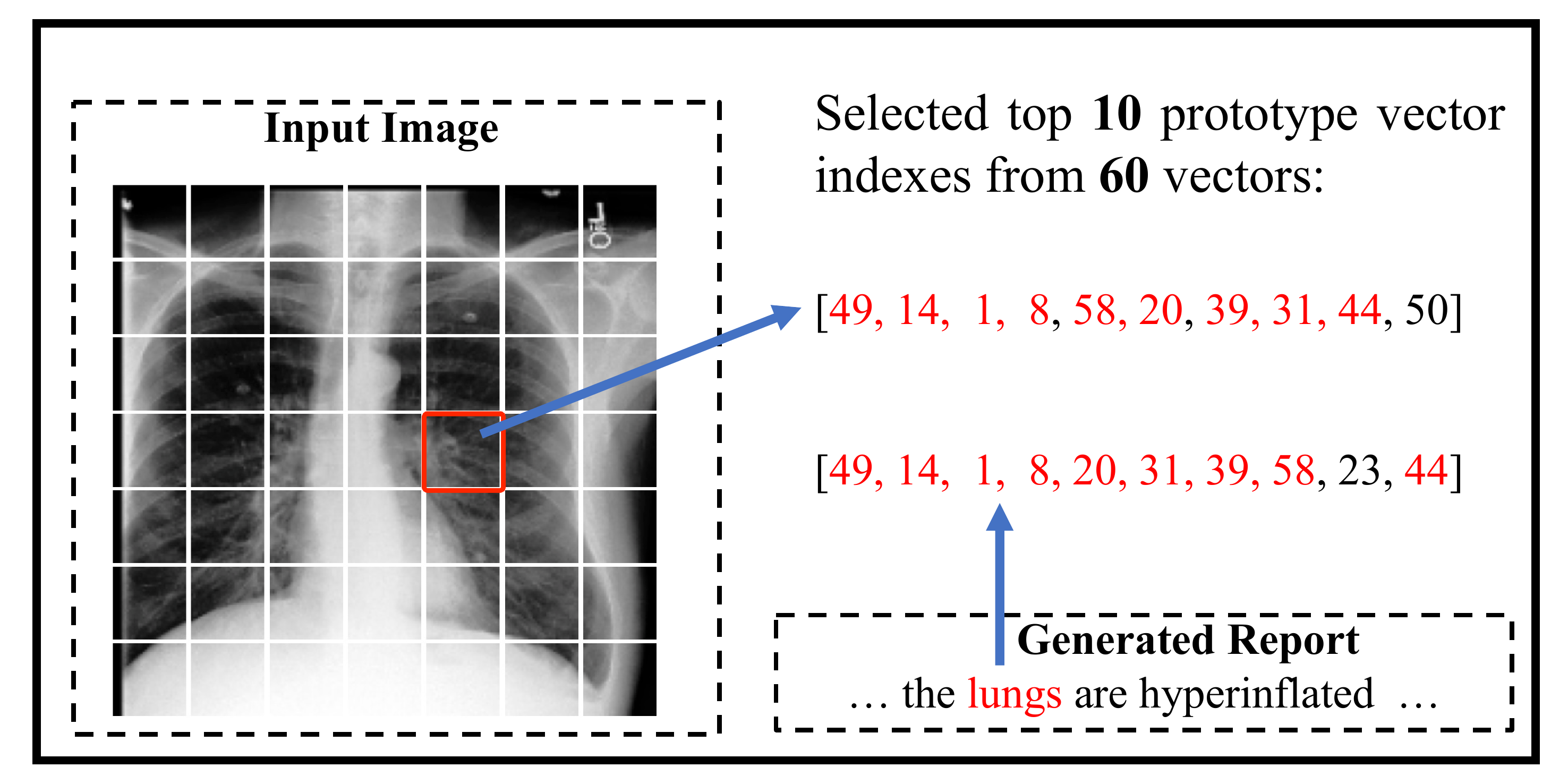}
  \captionsetup{width=.85\linewidth}
  \captionof{figure}{An example generated report and the selected cross-modal prototype indices using XPRONET. The selected word ``lungs'' is marked as red and the associated image patch is highlighted in the red rectangle. The prototype indices selected both from the image patch and from the text instance are marked as red. }
  \label{fig:Index_Example}
\end{minipage}%
\begin{minipage}{.4\textwidth}
  \centering
  \includegraphics[width=.95\linewidth]{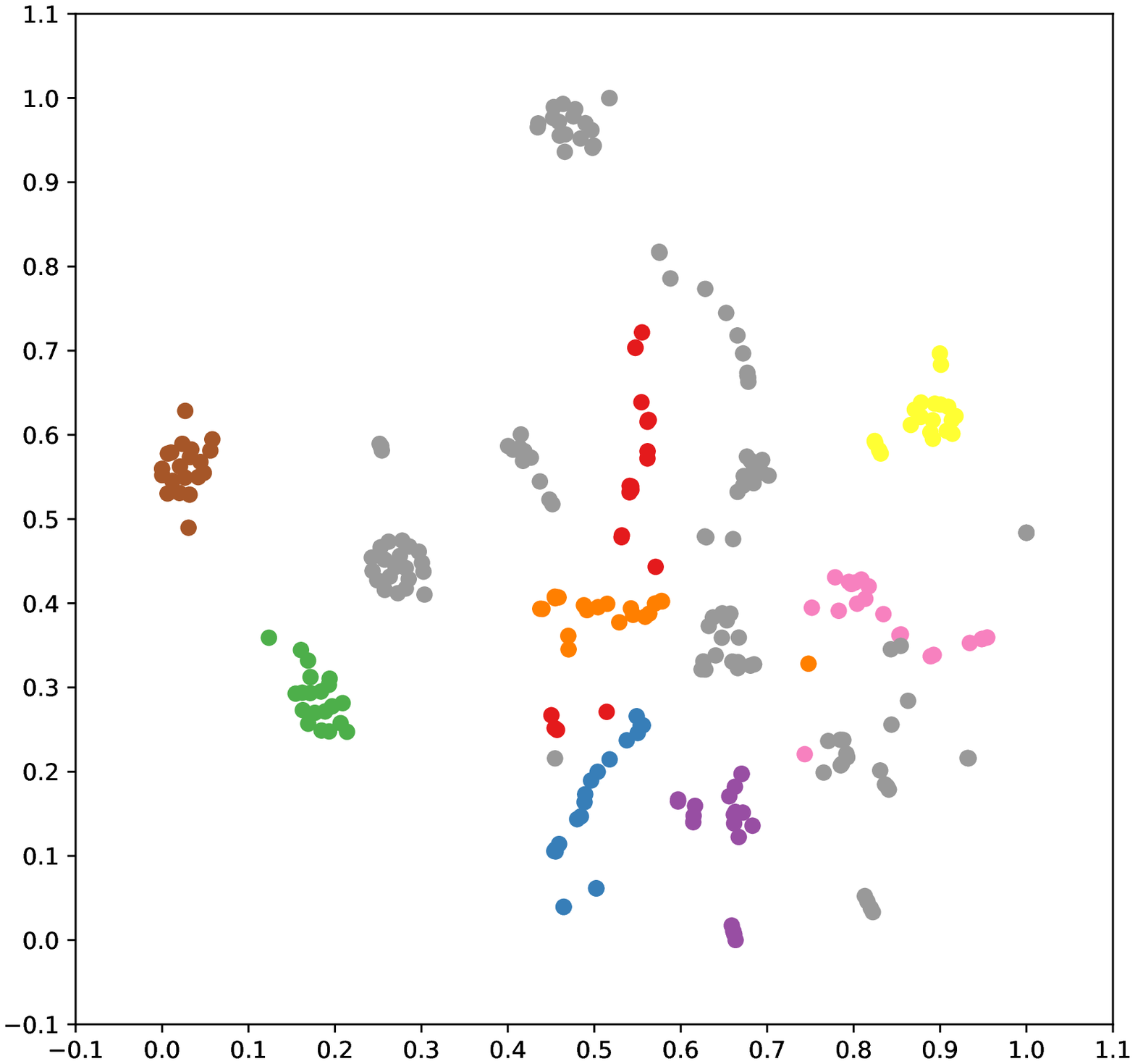}
  \captionsetup{width=.9\linewidth}
  \captionof{figure}{A visualization of the cross-modal prototype matrix on the MIMIC-CXR dataset using T-SNE~\cite{van2008visualizing} . Points with the same colour come from the same prototype category.}
  \label{fig:prototypes}
\end{minipage}
\end{figure}

Existing methods often focus on learning discriminative, {\em single}-modal features and ignore the importance of cross-modal interaction, essential for dealing with complex image and text semantic interrelationships. Thus, cross-modal interaction is of great importance as the model is required to generate a meaningful report only given the radiology image. Previous studies normally model cross-modal interaction by a self-attention mechanism on the extracted visual and textual features in an encoder-decoder architecture, which cannot adequately capture complex cross-modal patterns. Motivated by this, we propose a novel framework called \emph{Cross-modal PROtotype driven NETwork} (XPRONET) which learns the cross-modal prototypes on the fly and  utilizes them to embed cross-modal information into the single-model features. XPRONET regards the cross-modal prototypes as intermediate representations and explicitly establishes a cross-modal information flow to enrich  single-modal features.~\autoref{fig:Index_Example} shows an example of the cross-modal information flow where the visual and textual features select almost the same (9 of top 10) cross-modal prototypes to perform interaction. These enriched features are more likely to capture the sophisticated patterns required for accurate report generation. Additionally, the imbalance problem is addressed by forcing single-model features to interact with their cross-modal prototypes via a class-related, cross-modal prototype querying and responding module. Our work makes three principal contributions: 
\begin{enumerate}

\item We propose a novel end-to-end cross-modal prototype driven network where we utilize the cross-modal prototypes to enhance image and text pattern interactions. Leveraging cross-modal prototypes in this way for RRG has not been explicitly explored.   

\item We employ a memory matrix to learn and record the cross-modal prototypes which are regarded as intermediate representations between the visual and textual features. A cross-modal prototype network is designed to embed cross-modal information into the single-modal features. 

\item We propose an improved multi-label contrastive loss to learn cross-modal prototypes while simultaneously accommodating label differences via an adaptive controller term. 
\end{enumerate}
After a discussion of related work, our methods and implementation are described in detail in Section~\ref{sec-methods}. Experimental results presented in Section~\ref{sec-experiments} demonstrate that our approach outperforms a number of state-of-the-art methods over two widely-used benchmarks. We also undertake ablation studies to verify the effectiveness of individual components of our method. Discussion and proposals are given to inspire future work.

\section{Related Work}

\subsubsection{Image Captioning}
Image captioning aims to generate  human-like sentences to describe a given image. This task is considered as a high-level visual understanding problem which combines the research of computer vision and natural language processing. Recent state-of-the-art approaches~\cite{pei2019memory,wang2020multimodal,liu2018simnet,lu2017knowing,you2016image,rennie2017self} follow an encoder-decoder architecture and have demonstrated a great improvement in some traditional image captioning benchmarks. In particular, the most successful models~\cite{cornia2020meshed,guo2020normalized,pan2020x,ji2021improving} usually adopt the Transformer~\cite{vaswani2017attention} as their backbone due to its self-attention mechanism and its impressive capability of extracting meaningful features for the task. However, these methods are designed for short textual description generation and are less capable for generating long reports. Though several works~\cite{krause2017hierarchical,melas2018training} have been proposed to deal with long text generation, they often cannot capture the specific medical observations and tend to produce reports ignoring abnormal regions in images, resulting in unsatisfactory performance.   

\subsubsection{Radiology Report Generation}
Inspired by the great success of encoder-decoder based frameworks in image captioning, recent radiology report generation methods have also employed similar architectures. Specifically, Jing et al.~\cite{jing2018automatic} developed a hierarchical LSTM model to produce long reports and proposed a co-attention mechanism to detect abnormal patches. Liu et al.~\cite{liu2019clinically} proposed to firstly determine the topics of each report, which are then conditioned upon for report generation. Similarly, Zhang et al.~\cite{zhang2020radiology} also ascertained the disease topics and utilized prior knowledge to assist report generation via a pre-constructed knowledge graph. Liu et al.~\cite{liu2021exploring} extended this work by presenting a PPKED model which distills both the prior and posterior knowledge into report generation. A few works~\cite{miura2021improving,nishino2020reinforcement} have investigated reinforcement learning for improving the consistency of the generated reports. These encoder-decoder approaches often focus on extracting discriminative single-modal features (visual {\em{or}} textual), 
while few study explores the importance of the cross-modal pattern interactions. 

The most similar work to ours is R2GenCMN~\cite{chen2021cross} which utilizes an extra memory to learn the cross-modal patterns. Nonetheless, there are three main differences. First, we design a shared cross-modal prototype matrix to learn the class-related cross-modal patterns and propose an improved multi-label contrastive loss, while~Chen et al. \cite{chen2021cross} randomly initialize a memory matrix and use a cross entropy loss. Additionally, our querying and responding process is class-related, that is, cross-modal pattern learning is only performed over the cross-modal prototypes sharing the {\em same} labels rather than on all cross-modal prototypes. Moreover, we adopt a more effective approach to distill the cross-modal information into the single-modal representations rather than the simple averaging function used in R2GenCMN. XPRONET is driven by the cross-modal {\em prototypes} which, to the best of our knowledge, has not been explored before in radiology report generation. 

\section{Methods}
\label{sec-methods}

Our aim is to learn important informative cross-modal patterns and utilize them to explicitly model cross-modal feature interactions for radiology report generation,~\autoref{fig:architecture} shows the overall architecture of XPRONET. The details of the main three modules, i.e., the image feature extractor, the cross-modal prototype network, and the encoder-decoder are described in the following subsections.
\begin{figure*}[t]
    \centering 
    \includegraphics[width=0.95\textwidth]{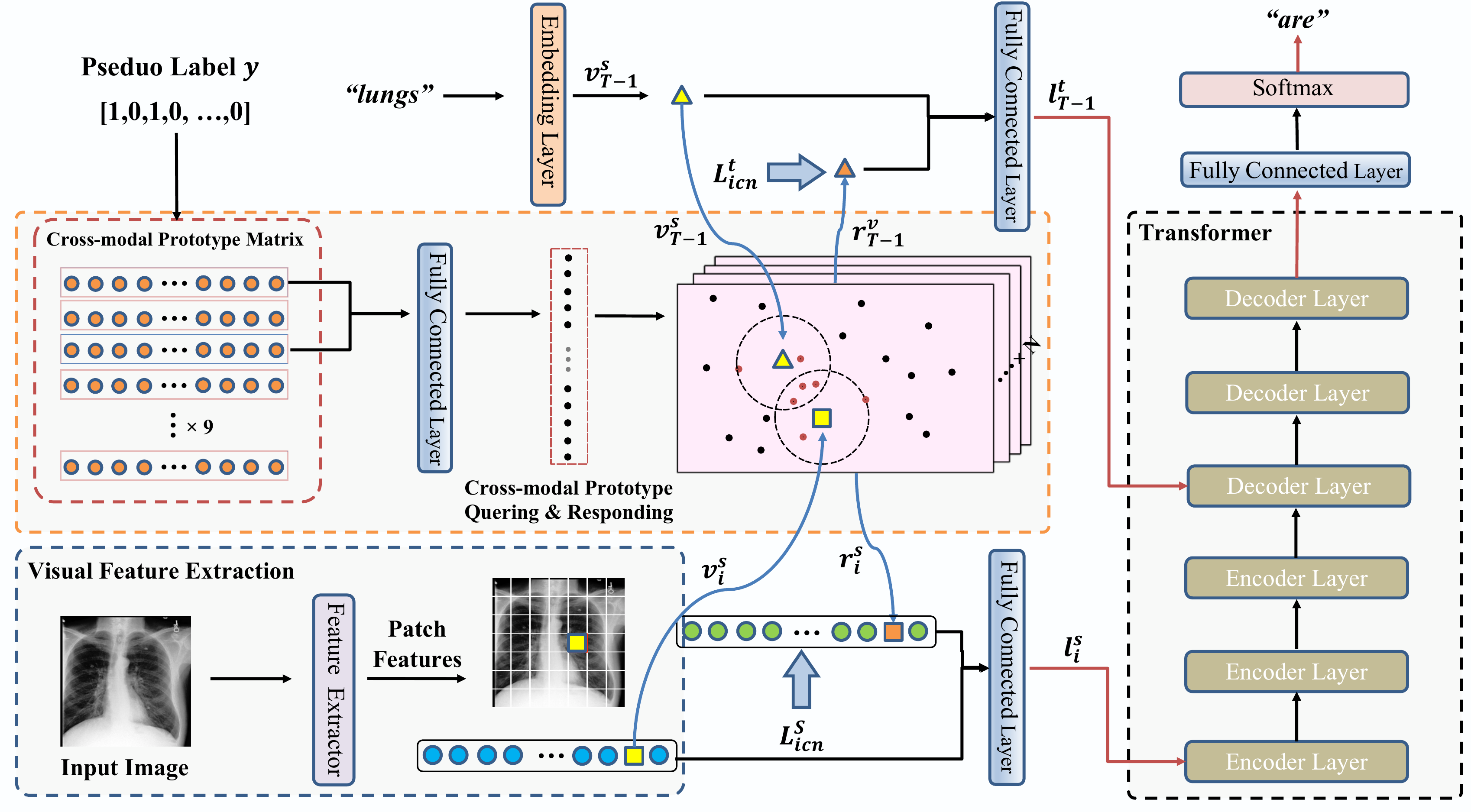} 
    \caption{The architecture of XPRONET: An image is fed into the Visual Feature Extractor to obtain patch features. A word at time step $T$ (e.g. ``lungs'') is mapped onto a word embedding via an embedding layer. The visual and textual representations are then sent to the cross-modal prototype querying and responding module to perform cross-modal interaction on the selected cross-modal prototypes based on the associated pseudo label. Then the single-model feature are enriched by the generated responses through a linear layer and taken as the source inputs of the Transformer encoder-decoder to generate the report.}    
    \label{fig:architecture}  
\end{figure*}

\subsection{Image Feature Extractor}
Given an input radiology image $\bm{I}$, a ResNet-101~\cite{he2016deep} is utilized to extract the image features $\bm{v}\in \mathbb{R}^{ H \times W \times C}$,  shown in the blue-dashed rectangle in ~\autoref{fig:architecture}. In particular, image features $\bm{v}$ are extracted from the last convolution layer, before the final average pooling operation. Here ${H,\ W \ and \ C}$ are the height, width and the number of channels of an image, respectively. Once extracted, we linearize the image features $\bm{v}$ by concatenating the rows of the image features and regard each region (position) feature as a visual word token. The final feature representation sequence $\bm{v}_s \in \mathbb{R}^{ HW \times C}$ is taken as the input for the subsequent modules and is expressed as:
\begin{equation}
 \{ v^s_1, v^s_2, ...,v^s_i, ..., v^s_{N^s-1}, v^s_{N^s} \} = f_{ife}(\bm{I}),
\end{equation}
where $v_i$ denotes the region features in the $i^{th}$ position of $\bm{v}_s$, $N^s=H \times W$, and $f_{ife}(\cdot)$ is the image feature extractor.

\subsection{Cross-modal Prototype Network}
Learning complex related patterns between image features and related textual descriptions is challenging. 
But cross-modal learning enables jointly learn informative representations of image {\em and} text. Central to our network is a prototype matrix which contains image pseudo-labels, initialized using an approach described below. 

\subsubsection{Pseudo Label Generation}
Cross-modal prototypes require category information for each sample, which is however often not provided in the datasets. To address this problem for prototype learning, we utilize CheXbert~\cite{smit2020combining}, an automatic radiology report labeler, to generate a pseudo label for each image-text pair. We denote the report associated with image $\bm{I}$ as:
\begin{equation}
\bm{R} = \{w_1, w_2, ..., w_i, ..., w_{N^r-1}, w_{N^r}\},
\end{equation}
where $w_i$ is the $i^{th}$ word in the report and $N^r$ is the number of words in the report. The labelling process can then be formulated as:
\begin{equation}
\{y_1, y_2, ..., y_i, ..., y_{N^l-1}, y_{N^l}\} = f_{al}(\bm{R}),
\end{equation}
where the result is an one-hot vector and $y_i \in \mathbb\{0, 1\}$ is the prediction result for $i^{th}$ category. Note that the value of one indicates the existence of that category, $N^l$ is the number of categories, and $f_{al}(\cdot)$ denotes the automatic radiology report labeler.

\subsubsection{Prototype Matrix Initialization}

Existing methods often directly model the cross-modal information interactions using the encoded features and learn implicitly cross-modal patterns. The length of the report, the imbalanced distribution of text descriptions of normal and abnormal cases, 
and complex cross modal patterns, make it hard to capture cross-modal patterns effectively. For better cross-modal pattern learning, we design a shared cross-modal prototype matrix $\bm{PM}\in\mathbb{R}^{N^l \times N^p \times D}$ to learn and store the cross-modal patterns, which can be considered as intermediate representations. Here $N^p$ and $D$ are the number of learned cross-modal prototypes for each category and the dimension for each prototype, respectively. $\bm{PM}$ is updated and learned during training, and then utilized by the class-related prototype querying and responding modules to explicitly embed the cross-modal information to the single-modal features. 

The initialization of the prototype matrix is critical. One way is to randomly initialize the matrix~\cite{chen2021cross}, but this does not capture any meaningful semantic information and hampers the subsequent prototype learning. Therefore, we propose to utilize prior information to initialize a semantic cross-modal prototype matrix. Specifically, for an image-text pair $<\bm{I},\bm{R}>$ with the associated pseudo class labels $\bm{y}$, we employ a pretrained ResNet-101 and BERT~\cite{smit2020combining} to extract the global visual and textual representations, $\bm{o}^i \in \mathbb{R}^{1 \times C_1}$ and $\bm{o}^t \in \mathbb{R}^{1 \times C_2}$, where $C_1$ and $C_2$ are the number of channels extracted of the visual and textual representations, respectively. To improve robustness, we also extract the flipped image features $\bm{o}^{if} \in \mathbb{R}^{1 \times C_1}$. By repeating this process on all the training samples, we can obtain a group of feature sets for each class, formulated as:
\begin{equation}
 \bm{R}^I_k = \{o^{i(f)}_u|y_{u,k} =1\}, \   \bm{R}^T_k = \{o^t_u|y_{u,k} =1\}.
\end{equation}
Here $\bm{R}^I_k$ and $\bm{R}^T_k$ are the visual and textual feature sets for category $k$, $i(f)$ means either the original image $i$ or the flipped image $if$, and $y_{u,k}$ denotes the label of category $k$ for sample $u$. After that, we concatenate the visual and textual representations to form the cross-modal features, $\bm{r} \in \mathbb{R}^{1 \times D}$. Note that $D= C_1 + C_2$. Finally, K-Means~\cite{lloyd1982least} is employed to cluster each feature set into $N_p$ clusters and the average of features in each cluster is used as an initial cross-modal prototype for $\bm{PM}$. This process can be summarised as:
\begin{gather}
 o_u = Concat(o^{i(f)}_u, o^t_u), \label{clustering}\\
 \{\bm{g}^k_1, ..., \bm{g}^k_{N^p-1}, \bm{g}^k_{N^p}\} = f_{km}(\bm{R}_k), \qquad 
 \bm{g}^k_i = \{o^{k,i}_1, ..., o^{k,i}_{N^d_{k,i}}\},
 \label{eqn-prototypes}\\
 \bm{PM}(k, i) =\frac{1}{N^s_{k,i}}{\sum_{j=0}^{N}r^{k,i}_j},
\end{gather}
Where $o_u$ and $\bm{R}_k$ are the concatenated cross-modal representation for sample $u$ and the cross-modal feature set for category $k$, $\bm{g}^k_i$ is the $i^{th} $ grouped cluster for $k^{th}$ category returned by the K-Mean algorithm $f_{km}$. ${N^d_{k,i}}$ is the number of samples in the $i^{th}$ cluster for $k^{th}$ category. $\bm{PM}(k,i)$ then represents the $i^{th}$ vector in the cross-modal prototype set for the $k^{th}$ category.

\subsubsection{Cross-modal Prototype Querying}
After obtaining the prototype matrix, similar to~\cite{chen2021cross}, we adopt a querying and responding process to explicitly embed the cross-modal information into the single-modal features. 
Different from~\cite{chen2021cross}, given an image, our cross-modal prototype querying measures the similarity between its single-modal representation and the cross-modal prototype vectors under the same label as the image, and selects the top $\gamma$ vectors having the highest similarity to interact with the single-model representations. This process is illustrated in the yellow-dashed rectangle in ~\autoref{fig:architecture}.

Given the image-text training pair $<\bm{I},\bm{R}>$ and the associated pseudo label $\bm{y}$, the queried cross-modal prototype vectors for the sample are then generated. 
The queried cross-modal prototype vectors $\bm{pv} = \{\bm{PM}(k)|\ y_k =1\} $, where $\bm{PM}(k)$ is the cross-modal prototype set for the $k^{th}$ category generated by Equations (5) - (8). To filter out possible noise, a linear projection is applied to $\bm{pv}$ to map it to $C_P$ dimensions before sending it into the querying process, as follows:
\begin{equation}
\label{eqn-querying}
\bm{p}=   \bm{pv} \cdot \bm{W}_{pv},
\end{equation}
\noindent where $W_{pv} \in \mathbb{R}^{D \times C_P}$ is a learnable weight matrix. 
\par
We denote the report representation output by the embedding layer as $\bm{v}_t = \{v^t_1, v^t_2,...,v^t_i,...,v^t_{N^t-1}, v^t_{N^t}\}$ and a cross-modal prototype vector as $p_i$, where $v^t_i \in \mathbb{R}^{1 \times C}$ is the $i^{th}$ word embedding of the report. Before performing the querying, we linearly project the visual feature sequence $\bm{v}_s$, textual report embeddings $\bm{v}_t$ and the cross-modal prototype vector $p_i$ into the same dimension $d$ since they may have different dimensions:
\begin{equation}
 v^{s*}_i =  v^s_i \cdot \bm{W}_v, \ \ \ v^{t*}_i =  v^t_i \cdot \bm{W}_v, \ \ \ p^*_i =   p_i \cdot \bm{W}_p,
\end{equation}
where $\bm{W}_v \in \mathbb{R}^{C \times d}$ and $\bm{W}_p \in \mathbb{R}^{C_P \times d}$ are two learnable weights. A similarity between each single-modal feature and cross-modal prototype vector pair is computed by:
\begin{equation}
 D^s_{(i,u)} = \frac{v^{s*}_j \cdot p^*_u}{d}, \ \ \ D^t_{(j,u)} = \frac{v^{t*}_j \cdot p^*_u}{d}.
\end{equation}
Since the majority of the cross-modal prototypes might be irrelevant to the queried vectors, which may introduce noisy cross-modal patterns, we only select $\gamma$ most similar vectors to respond to the query vectors. After that, we calculate the weights among these selected prototype vectors based on the similarities. This process between a cross-modal prototype $p^*_u$, a visual region representation $v^{s*}_i$ and a textual word embedding $v^{t*}_i$ is captured by: 
\begin{equation}
 w^s_{(i,u)} = \frac{D^s_{(i,u)}}{{\sum_{j=1}^{\gamma}D^s_{(i,j)}}}, \  \ \   w^t_{(i,u)} = \frac{D^t_{(i,u)}}{{\sum_{j=1}^{\gamma}D^t_{(i,j)}}}
\end{equation}

\subsubsection{Cross-modal Prototype Responding}
After obtaining the top $\gamma$ similar cross-modal prototype vectors and their weights, the next step is to generate the responses for the visual and textual features. In particular, we firstly transform the queried prototype vectors to the same representation space of the query vectors via a fully connected layer. The responses for the visual and textual features are created by taking the weighted sum over these transformed cross-modal prototype vectors:
\begin{align}
 &e^s_{(i,j)}=p^{s*}_{(i,j)} \cdot \bm{W}_e, &  &e^t_{(i,j)} =p^{t*}_{(i,j)} \cdot \bm{W}_e, \\
 &r^s_i= \sum_{j=1}^{\gamma} w^s_{(i,j)} \cdot e^s_{(i,j)}, &  &r^t_i =\sum_{j=1}^{\gamma} w^t_{(i,j)} \cdot e^t_{(i,j)},
\end{align}
where $ p^{s*}_{(i,j)}$ and $p^{t*}_{(i,j)}$ are the $j^{th}$ prototype vectors in the most similar cross-modal prototype sets for the $i^{th}$ image patch and word, respectively. Similarly, the $j^{th}$ transformed prototype vectors for $i^{th}$ image patch and word are denoted as $e^s_{(i,j)}$ and $e^t_{(i,j)}$. 
We represent the responses for $i^{th}$ image patch and word as $r^s_i$ and $r^t_i$. The $w^s_{(i,j)}$ and $w^t_{(i,j)}$ are the weights obtained by Equations (11) to (12).

\subsubsection{Feature Interaction Module}
The selected cross-modal prototype vectors contain class-related and cross-modal patterns. The last step is to introduce these informative patterns into the single-modal features via feature interaction. In~\cite{chen2021cross}, this is achieved by directly adding the single-model features and their associated responses, which pays the same attention to the responses and the single-modal features. However, this simple approach might be suboptimal given possibly noisy responses or non-discriminative single-model features. To mitigate this problem, we propose to automatically learn the importance difference and filter out noisy signals.

Specifically, we firstly concatenate the single-modal features with their associated responses. A linear layer is then applied to fuse the single-modal features and the cross-modal prototype vectors. Remember that the fused representations contain rich class-related features and cross-modal patterns. The process is:
\begin{equation}
  \bm{l}^s = \bm{FCN}(Concat(\bm{v}^s, \bm{r}^s)), \ \ \ \bm{l}^t = \bm{FCN}(Concat(\bm{v}^t, \bm{r}^t)),
\end{equation}
where $\bm{FCN}$ denotes the fully connected layer and $Concat$ is the concatenating function. The outputs of the Feature Interaction Module are taken as the source inputs for the following Transformer module to generate the reports.

\subsection{Reports Generation with Transformer}
Transformers have been shown to be quite potent for NLP tasks, e.g., sentiment analysis~\cite{wang2020transmodality,cheng2021multimodal,yang2020cm}, machine translation~\cite{zhang2018improving,wang2019learning,bao2021g} and question answering~\cite{naseem2021semantics,kacupaj2021conversational,zhao2020condition}. Consequently, we adopt a transformer to generate the final reports. Generally, the Transformer consists of the Encoder and Decoder. At the first step, the responded visual features $\bm{l}^s$ are fed into the Encoder to obtain the intermediate representations. Combined with the current fused textual representation sequence $\bm{l}^t = \{l^t_1, l^t_2, ..., l^t_i,..., l^s_{T-1}\}$ , these intermediate representations are then taken as the source inputs for the Decoder to predict the current output. In general, the encoding and decoding processes can be expressed as:
\begin{gather}
 \{m_1, m_2, ...,  m_{N^s}\} = \bm{Encoder}(l^s_1, l^s_2,..., l^s_{N^s}),\\
 p_T = \bm{Decoder}(m_1, m_2, ..., m_{N^s};l^t_1, l^t_2, ..., l^t_{T-1} ),
\end{gather}
where $p_T$ denotes the word prediction for time step $T$. The complete report is obtained by repeating the above process.

\subsection{Improved Multi-Label Contrastive Loss}  
Though the cross-modal prototype matrix is determistically initialized, further learning is required to learn class-related and informative cross-modal patterns, since the cross-modal patterns are actually far more sophisticated than the simple concatenation of the visual and textual representations in the Prototype Initialization module. Moreover, the cross-modal prototype features extractor (pre-trained ResNet-101 and BERT) are not trained on our target benchmarks, leading to potentially noisy signals. Therefore, online cross-modal prototype learning becomes of greater significance. 

A simple way is to utilize the widely used contrastive loss to supervise the learning of the cross-modal prototypes. Nonetheless, the vanilla contrastive loss is designed for the single-label prototype learning, while each training sample can belong to multiple categories in our task. Therefore, we modify the contrastive loss into a multi-label scenario by regarding the samples having at least one common label (excluding label 0) as positive pairs. If two samples do not share any common label, they form a negative pair. Instead of employing the contrastive loss on the responded features, we propose applying the loss on the responses since the fused features are used for medical report generation rather than for classification.

Given the visual responses $\bm{r}^s = \{r^s_1, r^s_2, ..., r^s_i,..., r^s_{N^s-1}, r^s_{N^s}\}$ and textual responses $\bm{r}^t = \{r^t_1, r^t_2, ..., r^t_i,..., r^t_{N^t-1}, r^s_{N^t}\}$, our modified multi-label contrastive loss is formulated as:
\begin{equation}
\label{eqn-ml-contrastive-loss}
\begin{split}
 \bm{L}^s_{icn} = \frac{1}{B^2} \sum_{i=1}^{B} 
 & \sum_{j:\bm{y}_i \otimes  \bm{y}_j \neq 0}^B  (\bm{\theta^{-\frac{h_d}{h_t}}}-Sim(\sigma(r^s_i, r^s_j))) +\\
 & \sum_{j:\bm{y}_i \otimes  \bm{y}_j = 0}^B  
    \max(Sim(\sigma(r^s_i, r^s_j)) - \alpha, 0) 
\end{split}
\end{equation}
Here $B$ denotes the number of training samples in one batch and $\otimes$ is the dot product operation. $\bm{y}_i \otimes \bm{y}_j \neq 0$ ensures that the responses $r^s_i$ and $r^s_j$ have at least one common label (excluding 0). $\sigma(\cdot)$ and $Sim(\cdot)$ are the the average function over all the image patch responses followed by the $L_2$ normalization and the cosine similarity function, respectively. Only negative pairs with similarity larger than a constant margin $\alpha$ can make a contribution to $\bm{L}^s_{cn}$. 

Note that different from a standard contrastive loss, the maximum positive similarity (or one) is replaced with a label difference term, $\theta^{(.)}$. In this way, the model can tolerate some dissimilarity between the positive pairs in terms of the label difference, instead of forcing them to be the same which is unreasonable under a multi-label setting: 
\begin{equation}
h_d = \epsilon(abs(\bm{y}_i - \bm{y}_j)), \ \ \ h_t = \epsilon(\bm{y}_i + \bm{y}_j),
\end{equation}
where $abs$ and $\epsilon$ are the absolute value and the summary functions, respectively. $h_d$ calculates the number of different labels and $h_t$ denotes the number of total labels of two training samples (excluding zero). Thus $\theta$ controls the relative tolerance where a smaller value represents less tolerance given the same label difference. An improved contrastive loss for textual responses $\bm{L}^t_{icn}$ is obtained in a similar way.

\subsubsection{Objective Function}
Given the entire predicted report sequence $\{ p_i \}$ and the associated ground truth report $\{ w_i \}$, XPRONET is jointly optimized with a cross-entropy loss and our improved multi-label contrastive loss by:
\begin{align}
\label{final_loss}
 \bm{L}_{ce} &= -\frac{1}{N^r}\sum_{i=1}^{N^r}w_i\cdot log(p_i), \\
\bm{L}_{fnl} &= \bm{L}_{ce} +\lambda \bm{L}^s_{icn} + \delta \bm{L}^t_{icn},
\end{align}
Here $\lambda$ an $\delta$ are two hyper-parameters which balance the loss contributions.

\begin{table}[t]
\caption{Comparative results of XPRONET with previous studies. The best values are highlighted in bold and the second best are underlined. BL, RG and MTOR are the abbreviations of BLEU, ROUGE and METEOR. The symbol $*$ denotes our replicated results with the official codes.}
\centering
\scriptsize
\label{tab:main_results}
\setlength\tabcolsep{3pt}
\begin{tabular}{clccccccc}
\toprule  
\textbf{Dataset} & \textbf{Method} & \textbf{BL-1} & \textbf{BL-2} & \textbf{BL-3} &\textbf{BL-4}  &\textbf{RG-L} & \textbf{MTOR} & \textbf{CIDEr} \\
\midrule

\multirow{9}{*}{\textbf{IU-Xray}} &$ST$~\cite{sukhbaatar2015end} &0.216 &0.124 &0.087 &0.066 &0.306 & - & -  \\
\multirow{9}{*} &$ADAATT$~\cite{lu2017knowing} &0.220 &0.127 &0.089 &0.068 &0.308 & - & 0.295\\
\multirow{9}{*} &$ATT2IN$~\cite{rennie2017self} &0.224 &0.129 &0.089 &0.068 &0.308 & - & 0.220 \\
\multirow{9}{*} &$SentSAT+KG$~\cite{zhang2020radiology} &0.441 &0.291 &0.203
&0.147 &0.304 & - & 0.304\\
\multirow{9}{*} &$HRGR$~\cite{li2018hybrid} &0.438 &0.298 &0.208 &0.151 &0.322 & - &\underline{0.343} \\
\multirow{9}{*} &$CoAT$\cite{jing2018automatic} &0.455 &0.288 &0.205 &0.154 &0.369 & - &0.277\\
\multirow{9}{*} &$CMAS-RL$~\cite{jing2019show} &0.464 &0.301 &0.210 &0.154 &0.362 & - & 0.275\\
\multirow{9}{*} &$KERP$~\cite{li2019knowledge} &\underline{0.482} &\underline{0.325} &\underline{0.226} &0.162 &0.339 & - &0.280\\
\multirow{9}{*} &$R2GenCMN^{*}$~\cite{chen2021cross} &0.474 &0.302 &0.220
&\underline{0.168} &\underline{0.370} & \underline{0.198} &-\\
\cline{2-9}
\multirow{9}{*}  &$\bm{XPRONET} (Ours)$  &\textbf{0.525} &\textbf{0.357} &\textbf{0.262} &\textbf{0.199} &\textbf{0.411} & \textbf{0.220} &\textbf{0.359}\\
\midrule

\multirow{8}{*}{\textbf{MIMIC}} &$RATCHET$~\cite{hou2021ratchet} &0.232 &- &- &- &0.240 & 0.101 &-\\
\multirow{8}{*}{\textbf{-CXR}} &$ST$~\cite{sukhbaatar2015end} &0.299 &0.184 &0.121 &0.084 &0.263 & 0.124 &-\\
\multirow{8}{*} &$ADAATT$~\cite{lu2017knowing} &0.299 &0.185 &0.124 &0.088 &0.266 & 0.118 &-\\
\multirow{8}{*} &$ATT2IN$~\cite{rennie2017self} &0.325 &0.203 &0.136 &0.096 &\underline{0.276} & 0.134 &- \\
\multirow{8}{*} &$TopDown$\cite{anderson2018bottom} &0.317 &0.195 &0.130 &0.092 &0.267 & 0.128  &-\\

\multirow{5}{*} &$R2GenCMN^{*}$~\cite{chen2021cross} &\textbf{0.354} &\underline{0.212} &\underline{0.139} &\underline{0.097} &0.271 & \underline{0.137} &-\\

\cline{2-9}

\multirow{5}{*}  &$\bm{XPRONET} (Ours)$  &\underline{0.344} &\textbf{0.215} &\textbf{0.146} &\textbf{0.105} &\textbf{0.279} & \textbf{0.138} &-\\

\bottomrule 
\end{tabular}
\end{table}

\section{Experiments}
\label{sec-experiments}
We verify the effectiveness of XPRONET on two widely used medical report generation benchmarks, i.e., IU-Xray and MIMIC-CXR. Four common natural language processing evaluation metrics:
BLEU\{1-4\}~\cite{papineni2002bleu}, ROUGE-L~\cite{lin2004rouge}, METEOR~\cite{denkowski2011meteor} and CIDEr~\cite{vedantam2015cider}, are utilized to gauge performance. The implementation details are given in Appendix A.1.

\subsubsection{Datasets}
IU-Xray~\cite{demner2016preparing} is a widely used benchmark which contains 7,470 X-ray images and 3,955 corresponding reports established by Indiana University. The majority of patients provided both the frontal and lateral radiology images. MIMIC-CXR~\cite{johnson2019mimic} is a recently released large chest X-ray dataset with 473,057 X-ray images and 206,563 reports provided by the Beth Israel Deaconess Medical Center. Both of these two datasets are publicly available~\footnote{https://openi.nlm.nih.gov/ \\ https://physionet.org/content/MIMIC-cxr-jpg/2.0.0/}. We follow the same data splits proportions as~\cite{li2018hybrid} to divide the IU-Xray dataset into train (70\%), validation (10\%) and test (20\%) sets, while the official data split is adopted for the MIMIC-CXR dataset.

\subsubsection{Comparisons with Previous Studies}
Here, we compare the experimental results with previous studies on the IU-Xray and MIMIC-CXR datasets. As shown in ~\autoref{tab:main_results}, ours (XPRONET) outperforms the previous best SOTA method of R2GenCMN by a noteable margin on the IU-Xray dataset. In particular, XPRONET surpasses the second best-performing method by 4.3\%, 3.1\% and 4.1\% on BLEU-1, BLEU-4 and RG-L scores respectively. A similar pattern can be seen on the MIMIC-CXR benchmark where XPRONET achieves the best performance on all the evaluation metrics except BLEU-1 in which it is slightly inferior to R2GenCMN. We mainly attribute the improved performance to the enriched single-modal feature representation via the cross-modal prototype learning. The superiority of XPRONET on IU-Xray is more obvious than MIMIC-CXR. This could be partly explained by the data size differences as the number of samples in MIMIC-CXR is almost 50 times larger than IU-Xray, hence it is more difficult to learn informative and class-related cross-modal prototypes. We present a visual example in ~\autoref{fig:example_vis} and give a further analysis below.

\subsubsection{Ablation Analysis}
Ablation studies were conducted to further explore the impact of each component of XPRONET on report generation performance. We investigated 
the following variants:
\begin{itemize}
\item \textbf{XPRONET w/o CMPNet}: the base model which only consists of the visual extractor (ResNet-101) and the encoder-decoder (Transformer) without other extensions.
\item \textbf{XPRONET w/o PI}: XPRONET without the cross-modal Prototype Initialization (PI), i.e., the cross-modal prototype matrix is randomly initialized.
\item \textbf{XPRONET w/o IMLCS}: XPRONET without the improved multi-label contrastive loss (IMLCS). We replace the adaptable maximum similarity $\theta^{-\frac{h_d}{h_t}}$ in Equation~(\ref{eqn-ml-contrastive-loss}) with one to switch it back to the standard multi-label contrastive loss.
\end{itemize}

\begin{table}[t]
\caption{The experimental results of ablation studies on the IU-Xray and MIMIC-CXR datasets. The best values are highlighted in bold. BL and RG are the abbreviations of BLEU and ROUGE.}
\centering
\label{tab:ablation_studies}
\setlength\tabcolsep{3pt}
\begin{tabular}{l|cccccc}
\toprule  
\textbf{IU-Xray}  & \textbf{BL-1} & \textbf{BL-2} & \textbf{BL-3} &
\textbf{BL-4}  &\textbf{RG-L} & \textbf{METEOR} \\
\midrule  

XPRONET  &\textbf{0.525} &\textbf{0.357} &\textbf{0.262} &\textbf{0.199} &\textbf{0.411} & \textbf{0.220} \\

w/o PI  &0.476 &0.307 &0.218 &0.160 &0.371 & 0.196 \\
w/o IMLCS  &0.471 &0.307 &0.215 &0.159 &0.377 & 0.196 \\
w/o CMPNet  &0.467 &0.303 &0.210 &0.155 &0.367 & 0.197 \\

\toprule  
\textbf{MIMIC-CXR}  & \textbf{BL-1} & \textbf{BL-2} & \textbf{BL-3} &
\textbf{BL-4}  &\textbf{RG-L} & \textbf{METEOR} \\
\midrule  

{XPRONET}  &\textbf{0.344} &\textbf{0.215} &\textbf{0.146} &\textbf{0.105} &\textbf{0.279} & \textbf{0.138}\\
w/o PI  &0.329 &0.205 &0.139 &0.100 &0.275 & 0.133 \\
w/o IMLCS  &0.336 &0.204 &0.137 &0.098 &0.269 & 0.135 \\
w/o CMPNet  &0.321 &0.198 &0.133 &0.095 &0.273 & 0.131 \\


\bottomrule 
\end{tabular}

\end{table}
\begin{figure}[t]
    \centering 
    \includegraphics[width=0.95\textwidth]{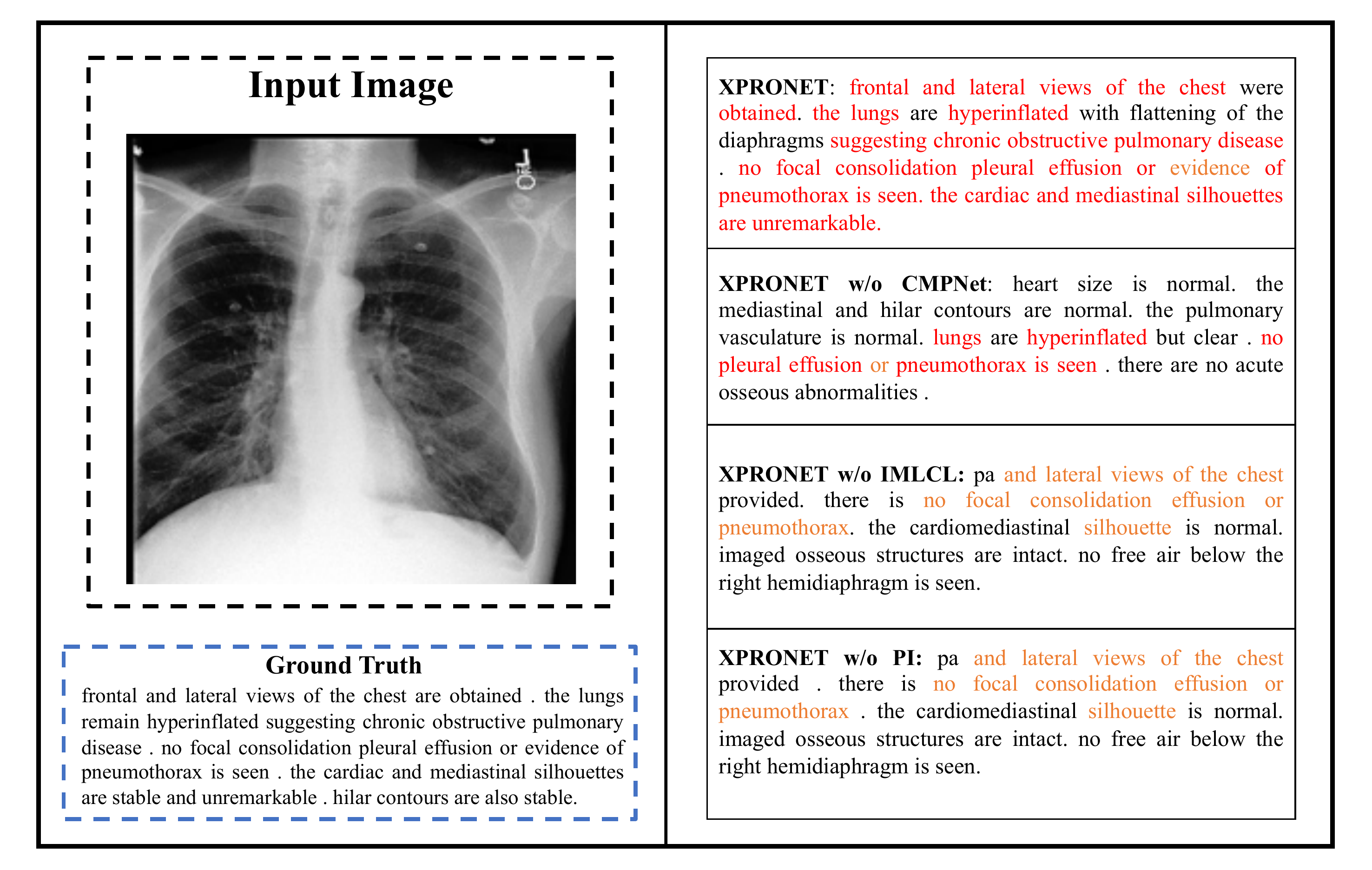} 
    \caption{An example of the report generated by different models. The ground truth report is shown in the blue dashed rectangle. Words that occurred in the ground truth are marked as red.  }    
    \label{fig:example_vis}  
\vspace{-15pt}
\end{figure}

The main results of the ablation studies of XPRONET are shown in~\autoref{tab:ablation_studies}. First, all the three components, i.e., prototype initialization, improved multi-label contrastive loss and the whole cross-modal prototype network architecture, significantly boost the performance as a notable drop can be seen when any of them is removed. For instance, the BLEU-4 score decreases from 0.199 to 0.160 and 0.105 to 0.100 on the IU-Xray and MIMIC-CXR datasets when the prototype initialization is removed. Similarly, removing the improved multi-label contrastive loss lead to 
lower scores on BLEU-2 and ROUGE-L. 
These results verify the importance of informatively initializing the cross-modal prototype and allowing some dissimilarity between positive pairs under the multi-label, cross-modal prototype learning settings. 

Moreover, the biggest performance drop can be seen on the model without the whole cross-modal prototype network on all the evaluation metrics, e.g., 0.525 to 0.467 and 0.344 to 0.321 of BL-1 on IU-Xray and MIMIC-CXR dataset respectively. An example visualization is shown in ~\autoref{fig:example_vis} to illustrate the strength of XPRONET.  More example visualizations are given in Appendix A.2. As we can see, XPRONET can capture the abnormal information and generate a better report, while the reaming models tend to produce sentences ignoring the abnormal patterns observed in images. This could be attributed to the well-learned cross-modal prototypes and the class-related querying and responding module which better capture the cross-modal flow and embed the prototype information into the feature learning procedure. We illustrate the cross-modal prototype matrix extracted from the linear projection (Equation~(\ref{eqn-querying})) in ~\autoref{fig:prototypes}. It can be seen that there is an obvious clustering pattern shown in the cross-modal prototype matrix. It should be mentioned that XPRONET can tolerate some dissimilarities between positive pairs, hence a category always occurring with other categories may lead to the associated prototypes being scattered with others (e.g., the orange), which is an expected outcome. To further explore the effectiveness of the XPRONET, we show an example of the generated report and the selected cross-modal prototype indices in ~\autoref{fig:Index_Example}. For the word ``lungs'' and its corresponding image patch, the majority (nine of ten) of their selected responding cross-modal prototypes are the same, indicating
that they learn the same cross-modal patterns and establish the cross-modal information flow via XPRONET, which is the expected behavior.

\begin{figure}
\centering
\begin{minipage}{.5\textwidth}
  \centering
  \includegraphics[width=.8\linewidth]{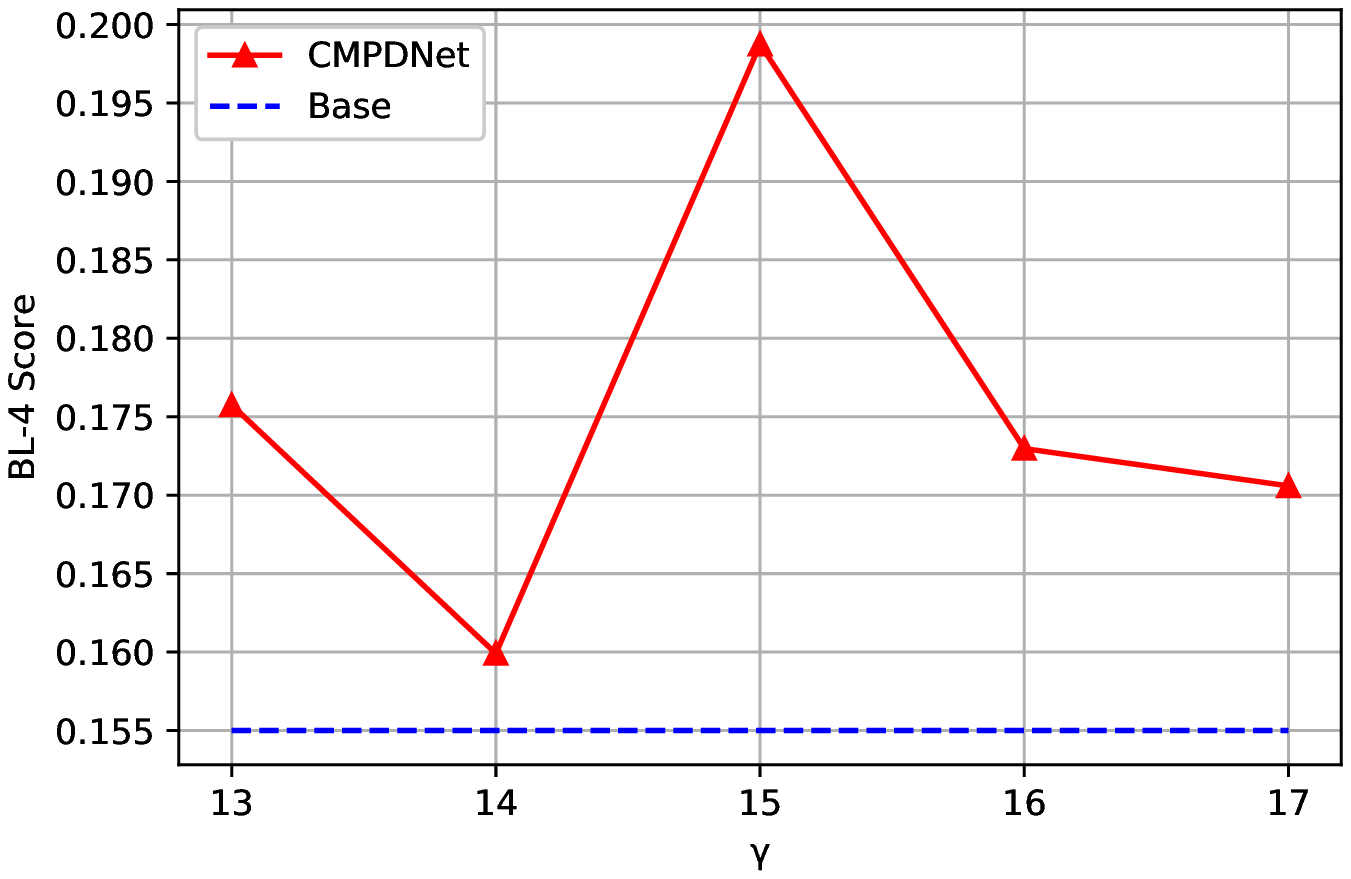}
  \captionsetup{width=.8\linewidth}
  \captionof{figure}{Effect of varying $\gamma$, number of responding prototype vectors on (BLEU-4 score).
  }
  \label{fig:gamma_ablation}
\end{minipage}%
\begin{minipage}{.5\textwidth}
  \centering
  \includegraphics[width=.8\linewidth]{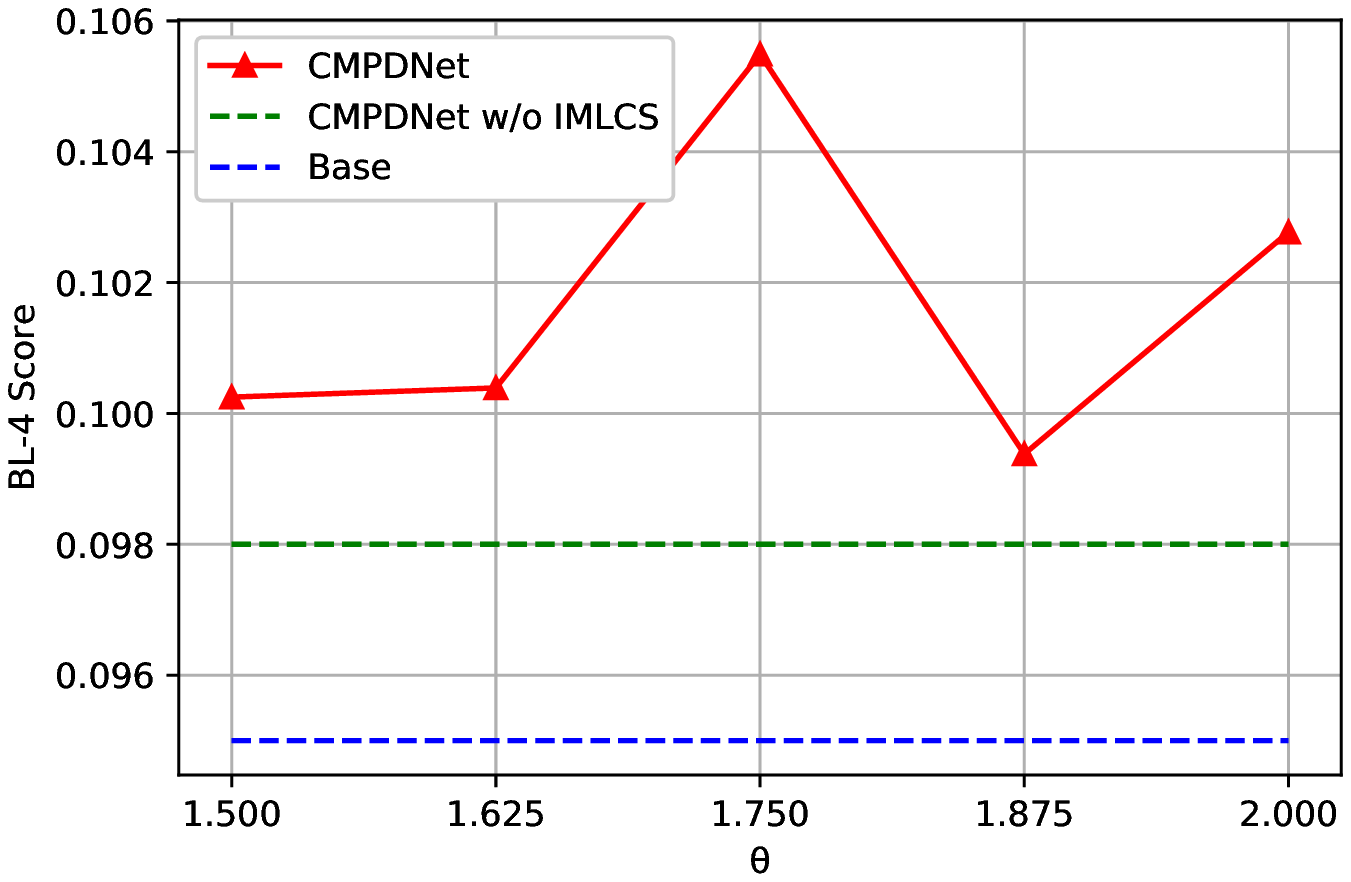}
  \captionsetup{width=.8\linewidth}
  \captionof{figure}{Effect of varying $\theta$, tolerance rate control on (BLEU-4 score).
  }
  \label{fig:theta_ablation}
\end{minipage}
\end{figure}

The sensitivity of XPRONET to the number of responding prototype vectors  $\gamma$ is shown in~\autoref{fig:gamma_ablation}. The BL-4 score reduces modestly when $\gamma$ increases from 13 to 14, and then culminates at $(0.199)$ at 15, after which the score decreases steadily to $0.171$ with the $\gamma$ increasing to 17 on the IU-Xray dataset. Generally, excessive or less responding prototype vectors can lead to notable performance drop. The reason being that excessive cross-modal prototype vectors may introduce noisy information, while inadequate numbers cannot provide sufficient cross-modal and class-related patterns. ~\autoref{fig:theta_ablation} illustrates the influence of the tolerance rate controller term $\theta$ of XPRONET on the MIMIC-CXR benchmark. As we can see, the best performance is achieved with a $\theta$ value of 1.750, and performance drops at other values. A smaller $\theta$ represents a larger maximum similarity which forces the positive pairs to be more similar, causing a performance drop given dissimilar positive pairs. In contrast, XPRONET cannot learn useful cross-modal prototypes with a large $\theta$ which leads to a small maximum similarity. Therefore, it appears important to strike a good balance between the cross-modal prototype learning and dissimilarity tolerance.

\section{Conclusions}

We propose a novel cross-modal prototype driven framework for medical report generation, XPRONET, which aims to explicitly model cross-modal pattern learning via a cross-modal prototype network. The class-related cross-modal prototype querying and responding module distills the cross-modal information into the single-model features and addresses the data bias problem. An improved multi-label contrastive loss is designed to better learn the cross-modal prototypes and can be easily incorporated into existing works. Experimental results on two publicly available benchmark datasets verify the superiority of XPRONET. We also provide ablation studies to demonstrate the effectiveness of the proposed component parts. A potential way to improve XPRONET is to increase the number of cross-modal prototypes, especially for larger datasets. In addition, we speculate that a more effective clustering approach in cross-modal prototype matrix initialization could bring further improvements.

\clearpage
%
%
\bibliographystyle{splncs04}
\bibliography{egbib}








\title{Cross-modal Prototype Driven Network for Radiology Report Generation\\
(Supplementary Material)} 

\author{Jun Wang, Abhir Bhalerao, Yulan He \\
Department of Computer Science, University of Warwick, UK}

\maketitle

\section{Implementation Details}

 Following the same strategy of previous work, e.g.~\cite{li2018hybrid,chen2021cross}, both images of a patient are utilized on IU-XRay and one image for MIMIC-CXR. In the training phase, images are first resized to $(256, 256)$ followed by a random cropping with the size of $(224, 224)$ before being fed into the model, while they are directly resized to $(224, 224)$ during the testing phase. We select the ResNet-101~\cite{he2016deep} pretrained on ImageNet~\cite{deng2009imagenet} as our visual extractor both in the prototype initialization module and our main task. Specifically, ResNet-101 produces patch features with $512$ dimensions for each one in the main task. In the prototype initialization module, ResNet-101 extracts global visual representation with $2048$ dimensions, and the global textual representation is obtain by a pretrained BERT~\cite{smit2020combining} with 768 dimensions.  

 We utilize a randomly initialized Transformer as the backbone for the encoder-decoder module with $3$ layers, $8$ attention heads and $512$ dimensions for the hidden states. The cross-modal prototype querying and responding follow a multi-head paradigm where each head has the same procedure as described in Section $3$. The number of clusters $N^P$ in equation $(6)$ is set to $20$. The pseudo label has $14$ categories, hence the cross-modal prototype matrix contain $14 \times 20=280$ vectors. $\gamma$ is set to $15$ which means we only select the top $15$ cross-modal prototype vectors to respond the single-modal representations. The term $\theta$ in the improved multi-label contrastive loss are $1.5$ and $1.75$ for the IU-Xray and MIMIC-CXR datasets respectively. 

 We use Adam as the optimizer~\cite{kingma2015adam} to optimize XPRONET under the cross entropy loss and our improved multi-label contrastive loss. $\lambda$ and $\epsilon$ in equation $(21)$ are $1$ and $0.1$. The learning rates are set to $1e-3$ and $2e-3$ for the visual extractor and encoder-decoder on IU-Xray, while MIMIC-CXR has a smaller learning rate with $5e-5$ and $1e-4$ respectively. The learning rates are decayed by $0.8$ per epoch and the bath sizes are 16 for all the datasets. The same as most promising studies, we adopt a beam size of three in the report generation to balance the effectiveness and efficiency. Note that the optimal hyper-parameters are determined by estimating the models on the validation sets. We implement our model via the PyTorch~\cite{paszke2019pytorch} deep learning framework.

\section{More Example Visualizations}
This section demonstrates more visualization results predicted by XPRONet.
\begin{figure}
    \centering 
    \includegraphics[width=0.9\textwidth]{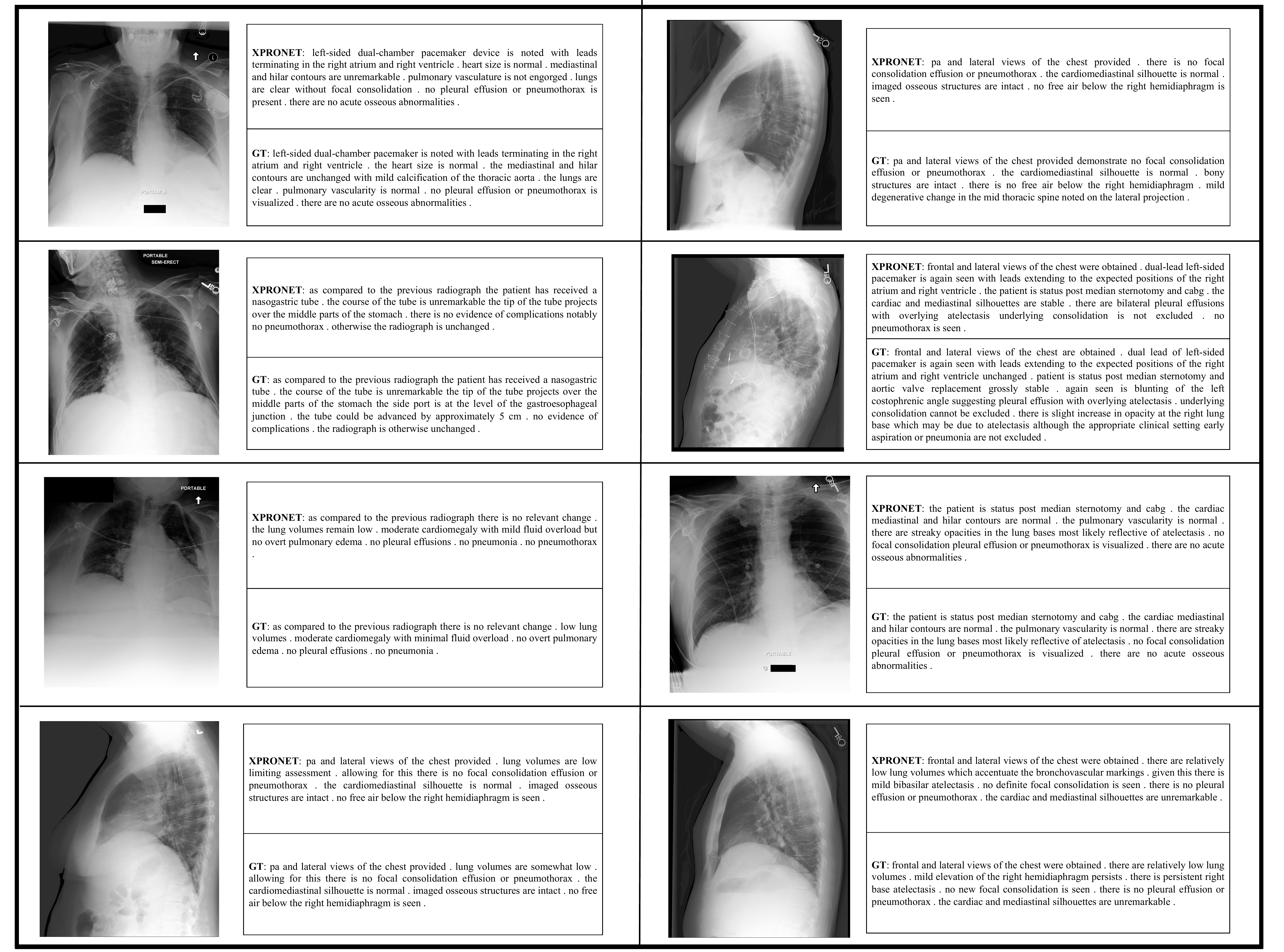} 
    \caption{The visulization of prediction results by XPRONET. GT is the abbreviation of the Ground Truth.}    
    \label{fig:architecture}  
\end{figure}

\clearpage



\end{document}